\documentclass[11pt,a4paper]{article}
\usepackage[hyperref]{styles/naaclhlt2019}
\usepackage{styles/private}
\usepackage{booktabs}
\usepackage{tikz}
\usepackage{verbatim}
\usetikzlibrary{shapes,arrows,positioning}

\aclfinalcopy 

\author{Yan Liang\textsuperscript{1},
  Xin Liu\textsuperscript{1},
  Jianwen Zhang\textsuperscript{2},
  and Yangqiu Song\textsuperscript{1} \\
  \textsuperscript{1}Department of CSE, Hong Kong University of Science and Technology, HK\\
  \textsuperscript{2}Microsoft, USA \\
  {\tt \textsuperscript{1}\{yliangav, xliucr, yqsong\}@cse.ust.hk } \\
  {\tt \textsuperscript{2}\{jiazhan\}@microsoft.com}\\}
\date{}

\title{Relation Discovery with Out-of-Relation Knowledge Base as Supervision}

\date{}

\begin{document}
\maketitle

\begin{abstract}

Unsupervised relation discovery aims to discover new relations from a given text corpus without annotated data. However, it does not consider existing human annotated knowledge bases even when they are relevant to the relations to be discovered. In this paper, we study the problem of how to use out-of-relation knowledge bases to supervise the discovery of unseen relations, where out-of-relation means that relations to discover from the text corpus and those in knowledge bases are not overlapped. We construct a set of constraints between entity pairs based on the knowledge base embedding and then incorporate constraints into the relation discovery by a variational auto-encoder based algorithm. Experiments show that our new approach can improve the state-of-the-art relation discovery performance by a large margin.

\end{abstract}


\section{Introduction}
\label{section:intro}

Relation extraction has been widely used for many applications, such as knowledge graph construction~\cite{GHHLMSSZ14}, information retrieval~\cite{LiuCFW14}, and question answering~\cite{RavichandranH02}.
Traditional supervised approaches require direct annotation on sentences with a relatively small number of relations~\cite{RothY02,Kambhatla04}.\footnote{We distinguish a relation (e.g., a predicate in a knowledge base) from the relation expression (e.g., the text surface between entities in a sentence) throughout the paper.}
With the development of large-scale knowledge bases (KBs) such as Freebase~\cite{freebase}, relation extraction has been extended to larger scales comparable to KBs using the distant supervision~\cite{mintz2009distant}.
However, when the training corpus does not support the annotated relations showing in the KB, such approach could fail to find sufficient training examples.
Distant supervision assumption can be violated by up to 31\% for some relations when aligning to NYT corpus \cite{RiedelYM10}.
More importantly, either traditional supervised learning or distantly supervised learning cannot discover new relations unseen in the training phase.

Unsupervised relation discovery tries to overcome the shortcomings of supervised or distantly supervised learning approaches.
Existing approaches either extract surface or syntactic patterns from sentences and use relation expressions as predicates (which result in many noisy relations)~\cite{knowitall,BankoCSBE07}, or cluster the relation expressions based on the extracted triplets to form relation clusters~\cite{YaoHRM11,YaoRM12,MarcheggianiT16}.
However, these approaches do not use existing high-quality and large-scale KBs when they are relevant to the relations to be discovered.

In this paper, we consider a new relation discovery problem where both the training corpus for relation clustering and a KB are available, but the relations in the training corpus and those in the KB are not overlapped.
As shown in \autoref{fig:illustration}, in the KB, we have entities \texttt{Pink Floyd}, \texttt{Animals}, etc., with some existing relations \textit{notable\_work} and \textit{has\_member} in the KB.
However, when doing relation discovery, we can only get supporting sentences that suggest new relations \textit{based\_on} and \textit{influenced\_by}.
This is a common and practical problem since predicates in KBs are limited to the annotator defined relations while the real relations in the world are always open and creative.

\begin{figure*}[t]
  \centering
\resizebox{.9\textwidth}{!}{
\begin{tikzpicture}[->,>=stealth',shorten >=1pt,auto,node distance=2.3cm,
  thick]

\tikzset{%
  base/.style = {rectangle, rounded corners, draw=black,
    text width=1cm, minimum height=0.5cm,
    text centered, font=\sffamily\tiny},
  main node/.style = {base},
  person/.style = {base, fill=blue!30},
  work/.style = {base, fill=green!30},
  sentence/.style = {text width=9cm, font=\sffamily\tiny}
}

  \node[person] (P) {Pink Floyd};
  \node[person] (R) [above=0.6cm of P] {Roger Waters};
  \node[work] (A) [right of=P] {Animals};
  \node[work] (AtD) [right of=R] {Amused  to Death};
  \node[work] (AOD) [right of=AtD] {Amusing Ourselves to Death};
  \node[work] (AF) [right of=A] {Animal Farm};
  \node[person] (G) [right of=AF] {George Orwell};
  \node[person] (N) [right of=AOD] {Neil Postman};

  \node[sentence] (sentence1) at (3.8,-1) { \textbf{Amused to Death}
    was inspried by Neil Postman's book \textbf{Amusing Ourselves to Death}.
  };
  \node[sentence] (sentence2) [below=0.1cm of sentence1] {Loosely based on George Orwell's  \textbf{Animal Farm}, \textbf{Animals} describe various classes in society as different kinds of animals
  };
  \node[sentence] (sentence2) [below=0.1cm of sentence2] {
\textbf{Postman} distinguishes the \textbf{Orwellian} vision of the future, from that offered by Aldous Huxley in Brave New World.
    };

  \tikzstyle{sfont}=[font=\sffamily\tiny,text width=1cm,
  align=center]
  \draw[->]  (P) -- node[sfont]{has member} (R)
  (P) -- node[sfont]{notable work} (A);
  \draw[->]  (R) -- node[sfont]{notable work} (AtD);
  \draw[->]  (N) -- node[sfont]{notable work} (AOD);
  \draw[->]  (G) -- node[sfont]{notable work} (AF);
  \draw[->,dotted]  (A) -- node[sfont]{based on} (AF);
  \draw[->,dotted]  (AtD) -- node[sfont]{based on} (AOD);
  \draw[->,dotted]  (N) -- node[sfont]{influenced by} (G);

\end{tikzpicture}
}
 \caption{An illustration of our new relation discovery setting.
The konwledge base contains relations \textit{notable\_work} and \textit{has\_member}.
However, the training corpus to perform relation discovery only contains new relations
\textit{based\_on} and \textit{influenced\_by}.}\label{fig:illustration}
\end{figure*}

It is challenging when there is no overlapped relation between target relation clusters and the KB because in this case the KB is not a direct supervision.
But if target relation clusters and the KB share some entities, we can use the shared entities as a bridge to introduce indirect supervision for the relation discovery problem.
Specifically, we build constraints between pairs of tuples based on the KB.
For example, in \autoref{fig:illustration}, when we cluster the \textit{based\_on} relation, we can evaluate the similarity between the tuple (\texttt{Animals}, \texttt{Animal Farm}) and the tuple (\texttt{Amused to Death}, \texttt{Amusing Ourselves to Death}) based on the KB.
If the KB tells us these two pairs of tuples are close to each other, then we put a constraint to force our relation clustering algorithm to group them together.

We use the discrete-state variational autoencoder (DVAE) framework~\cite{MarcheggianiT16} as our base relation discovery model since this framework is flexible to incorporate different features and currently the state-of-the-art.
We use KB embedding~\cite{BordesUGWY13} to obtain entity embeddings in the KB and use entity embeddings to evaluate the similarity between a pair of tuples.
Then constraints are constructed and incorporated into the DVAE framework in a way inspired by the must-link and cannot-link based constrained clustering~\cite{BasuBM04}.
We show that with no overlapped relations between the KB and the training corpus, we can improve the relation discovery by a large margin.

Our contributions are summarized as follows.

\begin{itemize}
\item We study a new prevalent but challenging task of relation discovery where the training corpus and the KB have no overlapped relation.
\item We propose a new kind of indirect supervision to relation discovery which is built based on pairwise constraints between two tuples.
  \item We show promising results using existing relation discovery datasets to demonstrate the effectiveness of our proposed learning algorithm for the new relation discovery task.
\end{itemize}

The code we used to train and evaluate our models is available at 
\url{https://github.com/HKUST-KnowComp/RE-RegDVAE}.


\section{Problem Definition}\label{section:problem}
We use $\ttrain$ to denote the set of all training sentences.
$\tentityset$ is the set of named entities that are recognized by an NER system in $\ttrain$, and $(e_1,e_2)$ is the  pair of first and second entities in a given sentence $x\in \ttrain$.
$\trelset$ is the set of relation labels for $\ttrain$.
In addition, there exists an external knowledge base $\KB$, consisting of a set of entities $\kentityset$ and relations $\krelset$ and triplets $\ktripletset$ where a triplet consists of two entities with their relation.

Our model is a relation extractor to predict the underlying semantic relation $r\in \trelset$ given sentences $\ttrain$, with the help of $\KB$.
In particular, we focus on the challenging scenario where $\trelset \cap \krelset = \emptyset$.

\section{Model}\label{sec:Model}
In this section, we first review the discrete-state variational autoencoder (DVAE) in \S\ref{section:VAE}.
Then we introduce our new framework in \S\ref{section:kbc}.

\subsection{DVAE for Relation Discovery}\label{section:VAE}
Assuming that we perform generative modeling, where each latent relation $r$ follows a uniform prior distribution $p_u(r)$, we follow \cite{MarcheggianiT16} to optimize a pseudo-likelihood:
\begin{align} \label{eq:vae.likelihood}
  \mathcal{L}(\theta) &= \log{\sum_{r\in \trelset} p(e_i,e_{-i}| r, \theta)p_u(r)}  \\
                      &\approx \sum_{i=1}^2\log{\sum_{r\in \trelset} p(e_i|e_{-i},r, \theta)p_u(r)},
\end{align}
where $e_i$ and $e_{-i}$ are entities, $i\in \{1, 2\}$ and $e_{-i}$ denotes the complement $\{e_1,e_2\} \setminus \{e_i\}$.
$p(e_i|e_{-i}, r, \theta)$ is the probability of one entity given another entity as well as the relation, where $\theta$ denotes the set of parameters.
Note that this probability $p$ is defined on the triplet $(e_1,r,e_2)$ which is universal across different sentences containing the two entities.

The pseudo-likelihood $\mathcal{L}(\theta)$ can be lower-bounded based on Jensen's inequality through a variational posterior $q(r|x, \psi)$:
\begin{align}\label{eq:vae.obj}
  \begin{split}
    \mathcal{L}(\theta, \psi) =&  \sum_{i=1}^2\sum_{r\in \mathcal{R}_T}q(r|x, \psi) \log{p(e_i|e_{-i},r,\theta)} \\
    & + \alpha H\left[q(r|x,\psi)\right],
  \end{split}
\end{align}
where $q(r|x, \psi)$ predicts the relation based on the whole sentence $x$ as an input
and $\psi$ as the set of parameters.
$H$ is the entropy to regularize the probability distribution $q$,
and $\alpha$ is the hyper-parameter to balance the regularization strength.

This model consists of two components, an encoder $q(r|x, \psi)$ which encodes sentence features into a relation distribution, and a decoder $p(e_i| r, e_{-i}, \theta)$ which predicts an entity given the relation cluster and another entity.
Both are modeled by softmax functions:
\begin{align}
  q(r|x,\psi) &=
                \softmax{\enwr \*g(x)}
                {\enwr[r']{} \*g(x)}
                {r'\in \trelset},  \label{eq:encoder} \\
  p(e_i|e_{-i},r, \theta) &=
                            \softmax{\decS{e_{-i}}}
                            {\decS[e_i']{e_{-i}}}
                            {e_{i}' \in \tentityset},
                            \label{eq:decoder}
\end{align}
where $\psi = \{\mathbf{w}_r | r \in \trelset\}$ and $\*g(x)$ is a vector representation of sentence $x$, which can be high-dimensional one-hot feature encodings or low-dimensional sentence embeddings encoded by deep neural networks. $\decS[e_1]{e_2}$ can be a general scoring function defined over triplets. We use the instantiation with the best performance shown by \cite{MarcheggianiT16}, which is a combination of bilinear model and selectional preference model:
\begin{align}
  \decS[e_1]{e_2} = \mathbf{e}_1^\transp {\bf C}_r \mathbf{e}_2
  + \left[ \mathbf{e}_{1},\mathbf{e}_{2}\right]^{\transp}\mathbf{r}
  \label{eq:decoder.score}
\end{align}
where $\theta = \{{\bf C}_r, \mathbf{r}, \mathbf{e}_i | r\in \mathcal{R}_T, e_i \in \mathcal{V} \}$,
${\bf C}_r$ is a matrix,
$\mathbf{r}$ is a vector for the relation $r$, $\mathbf{e}_{1}$ and $\mathbf{e}_{2}$ are the vectors for head and tail entities respectively,
and $\left[ \mathbf{e}_{1},\mathbf{e}_{2}\right]$
is the concatenation of the vector representations of the two entities.

The DVAE model directly optimizes the variational lower bound by doing gradient ascent for $\theta$ and $\psi$ jointly.
Both encoder $q(r|x, \psi)$ and decoder $p(e_i| r, e_{-i}, \theta)$ are implemented as neural networks. Standard training techniques and tricks can be applied.

\subsection{Knowledge Base Constraint}\label{section:kbc}
Our KB constraint framework can be summarized as a two-step procedure: KB constraints construction and regularization for the learning model.
In the constraints construction step,  a set of sentences is formed as a query to KB and retrieves a set of constraints back.
Then in the regularization step, we apply the constraint to regularize posterior distributions of the relation extractor.

Conceptually, given a set of sentences $\mathcal{X}$, we want to bias the learning result: After the entities are linked to the KB, if KB inference indicates that some pairs should be in a relation based on a set of rules $\Upsilon$, then the extractor should be constrained to output it.
This constraint can be encoded into a feature function $Q(\mathcal{X})$ = ``entity pairs in the same relation based on~$\Upsilon$'' and put into the posterior regularization framework~\cite{GillenwaterGGPT11}.
However, the computational complexity of the feature function is exponential since we need to traverse the KB to find $\Upsilon$.
We instead consider the must-link and cannot-link constraints~\cite{BasuBM04}, indicating respectively that a pair of sentences should be or should not be labeled as the same relation. 
For each pairwise constraint, the model assigns an associated cost of violating that constraint for the model regularization.

\begin{table*}[t]
  \small
  \centering
  \begin{tabular}{c|c}
    \toprule
    Euclidean $L_2$ distance & $d_{Euc}\left(\qr[1], \qr[2]\right) = \sqrt{ \sum_r{|\qr[1]-\qr[2]|}^2}$ \\
    \midrule
    Kullback-Leibler (KL) divergence & $d_{KL}( \qr[1], \qr[2]) = \sum_r \qr[1] \log \left(\frac{\qr[1]}{\qr[2]} \right)$ \\
    \midrule
    Jensen-Shannon (JS) divergence & $d_{JS}( \qr[1], \qr[2]) =       		\frac{1}{2}
                                     \sum_r \qr[1] \log \left(
                                     \frac{2\qr[1]}{\qr[1]+\qr[2]}
                                     \right)
                                     + \frac{1}{2}\sum_r \qr[2] \log\left(
                                     \frac{2\qr[2]}{\qr[1]+\qr[2]}
                                     \right)$ \\
    \bottomrule
  \end{tabular}
  \caption{Cluster regularization with different distance or divergences. } \label{tab:regularization}
  \vspace{-0.2in}
\end{table*}

\subsubsection{KB Constraints Construction} \label{sec:kbretrieval}

From the perspective of KB, a must-link constraint on sentences $(x_1, x_2)$ exists if two pairs of entities $(p_1, p_2) = [(e_{1,1},e_{1,2}),(e_{2,1}, e_{2,2})]$ are similar given the KB,
where $(e_{i,1}, e_{i,2})$ is the entity pair belongs to sentence $x_{i}$.
This motivates us to define a similarity score for a pair of entity pairs. Instead of modeling the common relation paths or logic rules, which is computationally infeasible, we compare them in the latent embedding space.
In particular, we model the KB using the TransE \cite{BordesUGWY13} model,  where a relation is interpreted as a translation from the head entity to the tail entity, with a score function, $\mathbf{e}_1 + \mathbf{r} = \mathbf{e}_2$  for each gold triplet $(e_1, r, e_2)$ in the KB.
This operation is fast and the latent embeddings are expressive in many cases. Then we can  reason the latent relation representation of a particular pair in vector space by $\mathbf{r}_i = \mathbf{e}_{i,2} - \mathbf{e}_{i,1}$, without the need for extra parameters.
Here $\mathbf{r}_i$ is not necessarily a real relation between two entities in the KB but just reflects the geometric property.
The penalty for violating a must-link constraint between a pair of sentences with a high KB score should be higher than those with low KB scores.
This further inspires us to define a soft constraint penalty based on the similarity of latent KB relations.

Here, we use the adjusted cosine similarity \cite{sarwar2001item} between two latent relations as a must-link confidence score
\begin{align}
  s^+(x_1, x_2) =  \left[\cos(\*e_{1,2} - \*e_{1,1}, \*e_{2,2}-\*e_{2,1})  \right]^+_{\gamma^+}
\end{align}
where $[x]^+_{\gamma^+}=x$ if $x>{\gamma^+}$ otherwise 0, ${\gamma^+} \in [0,1]$ is a threshold we defined to control the must-link scope, $e_{i,j}$ is named entity in $x_i$ and $\*e_{i,j}$ is its embedding.
The similarity between $\*e_{1,2} - \*e_{1,1}$ and $\*e_{2,2}-\*e_{2,1}$ evaluates whether two sentences indicate similar relations according to the KB embedding.

We also define the cannot-link in a similar way, where two sentences cannot be in the same cluster with a confidence
\begin{align}
  s^-(x_1, x_2) =  \left[\cos(\*e_{1,2} - \*e_{1,1}, \*e_{2,2}-\*e_{2,1})  \right]^-_{\gamma^-}
\end{align}
where $[x]^-_{\gamma^-}=x$ if $x<-{\gamma^-}$ otherwise 0, and ${\gamma^-} \in [0,1]$ is a threshold we defined to control the cannot-link scope.
We simply set ${\gamma^+} = {\gamma^-} = \gamma$.

\subsubsection{Clustering Regularization}

For each pair of sentences $(x_1, x_2)$, the relation extractor will predict a clustering posterior $q_i(r|x_i, \psi),\; i=1,2$, which can be computed based on Eq.~(\ref{eq:encoder}).
We regularize the clustering result on the probability distance between sentence pairs, using either Euclidean $L_2$ distance,  Kullback-Leibler (KL) divergence, or Jensen-Shannon (JS) divergence.
The computation of the distance or divergences can be found in Table~\ref{tab:regularization}.

Then the soft constraints introduced in \S\ref{sec:kbretrieval} are applied on the corresponding distance to calculate the regularization terms:
\begin{align}
  D^+(x_1, x_2) &= -d_*\left(\qr[1], \qr[2]\right)s^+(x_1, x_2), \\
  D^-(x_1, x_2) &= d_*\left(\qr[1], \qr[2]\right)|s^-(x_1, x_2)|,
\end{align}
for must and cannot links respectively, where $d_*$ can be $d_{Euc}$, $d_{KL}$, or $d_{JS}$.
Taking must-link constraint as an example, if the posterior distributions $q_1(r|x_1, \psi)$ and $q_2(r|x_2, \psi)$ are different from each other but KB suggests that these two sentences should be in the same cluster where $s^+(x_1, x_2)$ is large, then $d_*\left(\qr[1], \qr[2]\right)$ being large means there is a large cost when $q_1$ and $q_2$ being different.
Then in the training phase, we want to reduce this cost given the constraint.

The constraints above are defined in a $|\mathcal{X}| \times |\mathcal{X}|$ space.
It is almost impossible to enumerate all of the constraints.
To make it trainable, we instead gather the constraints  within a mini-batch.
Since in different training epochs we randomly permute the training samples, it is possible to touch many pairs of sentences in practice.

\subsection{Learning} \label{sec:learning}
The model parameters only exist in original autoencoder components (i.e., $\psi$ and $\theta$), which can be jointly optimized by maximizing the following objective function with $L_2$ regularization:
\begin{align}
  \begin{split}
    \mathcal{L}(\theta, \psi) =&  \sum_{x\in X}\sum_{i=1}^2\sum_{r\in R_T}q(r|x, \psi) \log{p(e_i|e_{-i},r,\theta)} \\
    & + \sum_{x\in X} \alpha H\left[q(r|x,\psi)\right] \\
    & + \sum_{X_i \sim X}\sum_{ (x_1,x_2) \in X_i }\beta D(x_1, x_2)\\
    & + \lambda \Vert (\psi, \theta) \Vert_2,
  \end{split} 	\label{eq:obj}
      \vspace{-0.25in}
\end{align}
where $\alpha$, $\beta$, $\gamma$, and $\lambda$  are hyper-parameters to control the regularization strength. $D$ can be $D^+$ or $D^-$ depending on the cosine similarity between pairs.  In practice, we apply annealing method over $\alpha$ in an exponential way:
\begin{align*}
  \alpha_t = \alpha_0 \exp(-\eta t) \;\textrm{and}\;
  \eta = \frac{\log(\alpha_0 / \alpha_T) }{T},
\end{align*}
where $\alpha_0$ is the initial value, and $\alpha_T$ is the final value, $t$ and $T$ are the current and total training steps respectively.
This method enables the extractor to explore more possibilities first and finally converge to a stable distribution.

It is difficult to directly compute the partition function in Eq.~(\ref{eq:decoder}), as it requires to sum over $|\mathcal{V}|$. We use the same negative sampling method as \cite{MarcheggianiT16} to substitute $\log{p(e_i|e_{-i},r,\theta)} $ in Eq.~(\ref{eq:obj}) with:
\begin{align*}
\log{p(e_i|e_{-i},r, \theta)}& \approx
   \log\sigma(  \decS[e_i]{e_{-i}}) \\
  +  \sum_{e^{\text{neg}}\in \mathcal{N}}&
     \log\sigma\left( -\decS[e^\text{neg}]{e_i}
     \right),
\end{align*}
where $\mathcal{N}$ is the set of randomly sampled entities in $\mathcal{V}$ and $\sigma$ is the sigmoid function.


\section{Experiments} \label{sec:exp}
In this section, we show the experimental results.
	\subsection{Dataset and Preprocessing}

	We evaluate our model in the context of unsupervised relation discovery and compare to the baseline model, DVAE \cite{MarcheggianiT16} which is the current state-of-the-art of relation discovery. Distant supervision assumes that the relations should be aligned between the KB and the training text corpus, which is not available in our setting.

    We tested our model on three different subsets of New York Times corpus (NYT) \cite{nyt}.
    \begin{itemize}
    \item The first one is widely used in unsupervised settings, which was developed by \citet{YaoHRM11} and has also been used by \citet{MarcheggianiT16}. This dataset contains articles  2000 to 2007, with named entities annotated and features processed (POS tagging, NER, and syntactic parsing). We use this dataset to compare with previous work directly \cite{MarcheggianiT16}.
      \item The second and third ones are usually applied by supervised models.
        So when they generated the data, they tended to focus on relations with more supporting sentences.
        The second one was developed by \citet{ZengLLS17}. The data is built by aligning Wikidata \cite{Vrandecic2012} relations with NYT corpus, as a result of 99 possible relations. It is built to contain more updated facts and richer structures of relations, e.g., a larger number of relation/relation paths. We use this dataset to amplify the effects coming from relation paths in KB, as the data was used to train a path-based relation extraction model.
    \item The third  one was developed by \citet{RiedelYM10} and has also been used by \citet{LinSLLS16}. This dataset was generated by aligning Freebase \cite{freebase} relations with NYT in 2005-2007, and with 52 possible relations. We use this data to test the clustering result with a narrow relation domain.
    \end{itemize}

\begin{table}[t]
    \centering
    {\small
        \begin{tabular}{ll|ccc}
    	\toprule
            \multicolumn{2}{c}{\textbf{Data}} &
            	\multicolumn{1}{|c}{\textbf{NYT122}} &
                \textbf{NYT71} &
                \textbf{NYT27} \\
            \midrule
            \multirow{ 5}{*}{\textbf{Text}}
			& \# sentences & 67,123 & 14,210 & 87,144 \\
			& \# facts & 9,207 & 2,274 & 8,559 \\
			& \# entity pairs & 20,939 & 3,539 & 36,714 \\
			& \# entities & 5,865 & 2,489 & 4,803 \\
			& \# relations & 122 & 71 & 27\\
			\midrule
			\multirow{4}{*}{\textbf{KB}}
			& \# triplets & 401,490 & 456,146 & 439,507\\
			& \# entity pairs & 331,008 & 373,875 & 354,960 \\
			& \# entities & 14,907 & 14,933 & 14,911\\
			& \# relations & 705 & 1,009  & 1,031  \\
            \bottomrule
        \end{tabular}
        }
\caption{Statistics of datasets. \# facts in the text corpus is the number of sentences with relation labels.}
    \label{table:dataset}
\end{table}
	We align these datasets against FB15K, which is a randomly sampled subset of Freebase developed by \citet{BordesUGWY13}. For each of the datasets above, we hold out the triplets in FB15K that contains  relations in corresponding text data, so that we ensure that KB cannot give any direct supervision on any relation labels. We then discard named entities in text corpus if they are not shown in KB, so that we can directly test the influence of our KB constraint model.
    Finally, we only keep a single label for each sentence, and $e_1$, $e_2$ follow the occurrence order in the sentence.
    The resulting datasets contain 122, 71, and 27 relation labels respectively, so we name them as
    NYT122,
    NYT71, and
    NYT27.
    The statistics of the three datasets are shown in Table \ref{table:dataset}.
    For NYT71 and NYT27, we perform the same feature extraction as NYT122 shown in \cite{MarcheggianiT16}.

    \subsection{Implementation Details}
    All the model parameters are initialized randomly.
    The number of negative samples is set to 5,
    mini-batch size is set to 100 with 80 epochs.
    We optimize all the models using AdaGrad \cite{DuchiHS11} with initial learning rate at 0.5.
    For NYT122, we induce 40 relations clusters, with
    	$ \alpha_0 = 4$, $\alpha_T = 10^{-5}$,
    	$\beta = 0.6$, and
        $\gamma = 0.9$.
    For NYT71, we induce 30 relations clusters, with
    	$ \alpha_0 = 2$, $\alpha_T = 10^{-4}$,
    	$\beta=0.8$, and
        $\gamma=0.95 $ .
	For NYT27, we induce 20 relations clusters, with
    	$ \alpha_0 = 2$, $\alpha_T = 10^{-4}$,
    	$\beta = 0.8$, and
        $\gamma = 0.3$.
    We train TransE as our KB embedding model with 50 dimensions and 1,000 epochs.

     We report the average and standard deviation based on five different runs. We randomly split the data into validation:test=4:6. All the model selections were based on validation sets, and final evaluation results will be only based on test sets.

    \subsection{Evaluation and Discussion}
    As the scoring function, we use the $B^3 F_1$ \cite{Bagga98} which has also been used by our baseline \cite{MarcheggianiT16}, and Normalized Mutual Information (NMI) \cite{StrehlG02} metrics. Both are standard measures for evaluating clustering tasks.

\begin{table*}[t]
\vspace{-0.1in}
    \centering
    {\footnotesize
        \begin{tabular}{l|cc|cc|cc|cc}
            \toprule
            \textbf{Model} & \multicolumn{8}{c}{\textbf{Metrics}} \\ \cline{2-9}
            & \multicolumn{4}{c}{Prediction based on encoder} & \multicolumn{4}{|c}{Prediction based on decoder} \\\cline{2-9}
            & \multicolumn{2}{c}{F1}  & \multicolumn{2}{|c}{NMI}  & \multicolumn{2}{|c}{F1}  & \multicolumn{2}{|c}{NMI} \\\cline{2-9}
           & Mean & Std & Mean & Std & Mean & Std & Mean & Std \\
            \midrule
            DVAE & 0.417 & 0.011 & 0.339 & 0.009 & 0.419 & 0.011 & 0.337 & 0.014 \\
            \midrule
            RegDVAE (Euclidean at encoder)& \textbf{0.469} & 0.014 & \textbf{0.430} & 0.020 & \textbf{0.448} & 0.020 & \textbf{0.384} & 0.020 \\

            RegDVAE (KL at encoder)& 0.375 & 0.009 & 0.359 & 0.014 & 0.380 & 0.011 & 0.355 & 0.014 \\

            RegDVAE (JS at encoder)& 0.435 & 0.038 & 0.370 & 0.042 & 0.409 & 0.012 & 0.336 & 0.005 \\ \midrule
            RegDVAE (Euclidean at decoder)  & 0.416 & 0.019 & 0.329 & 0.017 & 0.350 & 0.012 & 0.201 & 0.054\\
            \bottomrule

        \end{tabular}
        }
\vspace{-0.1in}
  \caption{Comparison results on NYT122 with different prediction and regularization strategies (using encoder or decoder).  }
 \label{table:modeltable}
\vspace{-0.1in}
\end{table*}

	\paragraph{Regularization and Prediction Strategies.}

    We first report our results on NYT122 using different regularization and prediction settings, as this dataset was used by our baseline model DVAE.

    Note that both encoder and decoder components can make relation predictions.
    In fact, the way of using encoder $q(r|x, \psi)$ for each sentence is straightforward.
    Then based on the encoder, we predict relation on the basis of single occurrence of entity pair.
    When using the decoder, we need to re-normalize $p(e_i| r, e_{-i}, \theta)$ as $p(r| e_1, e_2, \theta)$ to make predictions.
    Based on the decoder, we make predictions for each unique entity pair.
    As a consequence, our constraints can be imposed on both encoder and decoder.
    The way of computing decoder probability distribution is the same as making predictions.
    So in this experiment, we report both results.

    The results are shown in Table \ref{table:modeltable}.
    From the table, we can see that regularization with Euclidean distance performs the best compared to KL and JS.
    Moreover, the regularization over encoder is better than the regularization over decoder.
    This may be because the way that we put constraints only over sampled sentences in a batch may hurt the regularization of decoder, since sampled unique pairs may be less than sample sentences.
    If we look at results comparing original DVAE prediction based on the encoder and the decoder, both result in similar F1 and NMI numbers.
    Thus, we can only conclude that currently in the way we do sampling, constraining over encoder is a better choice.

	\paragraph{Comparison on Different Datasets.}
    We also compare our algorithm on the three datasets with different baseline settings.
    In order to evaluate our model rigorously, besides the original DVAE model, we compare two additional augmented baseline models with the same hyper-parameter setting: DVAE with TransE embeddings appended to encoder input features (DVAE+E) and DVAE with decoder entity vectors replaced by pre-trained KB embeddings (DVAE+D).
    For our method, we report RegDVAE with the best setting where we use Euclidean distance based constraints to regularize the encoder. Moreover, we report a setting with fixed embeddings in the decoder as the ones obtained from TransE (RegDVAE+D).
    This also makes sense since even though the TransE embeddings are not trained with the observation of the same relations as the text corpus, the embeddings already contain much semantic information about entities.
    Then by fixing the embeddings of entities in the decoder, we can significantly reduce the number of parameters that need to be trained.
    The results are shown in Table~\ref{table:comparetable}.
    As we can see that, RegDVAE+D can outperform the original DVAE by 8$\sim$23 points on F1.
    DVAE+D is also good but may fail when there are a lot of out-of-sample entities in the training corpus.

\begin{table*}[t]
    \centering
    {\footnotesize
        \begin{tabular}{l|cc|cc|cc|cc|cc|cc}
            \toprule
            \textbf{Model} & \multicolumn{4}{c}{\textbf{NYT122}} & \multicolumn{4}{|c}{\textbf{NYT71}} &
\multicolumn{4}{|c}{\textbf{NYT27}} \\\cline{2-13}
            & \multicolumn{2}{c}{F1}  & \multicolumn{2}{|c}{NMI} & \multicolumn{2}{|c}{F1}  & \multicolumn{2}{|c}{NMI} & \multicolumn{2}{|c}{F1}  & \multicolumn{2}{|c}{NMI} \\\cline{2-13}
            & Mean & Std & Mean & Std & Mean & Std & Mean & Std & Mean & Std & Mean & Std \\
            \midrule
            Majority & 0.355 & - & 0 & - &
            0.121 & - & 0 & - &
			0.549 & - & 0 & - \\ \cline{1-13}
			DVAE & 0.417 & 0.011 & 0.339 & 0.009 &
			0.325 & 0.011 & 0.375 & 0.023 &
			0.433 & 0.018 & 0.384 & 0.021 \\
			DVAE+E & 0.385 & 0.021 & 0.341 & 0.043 &
			0.339  & 0.021 & 0.418 & 0.022 &
			0.396 & 0.034 & 0.381 & 0.039 \\
			DVAE+D & 0.452 & 0.033 & 0.438 & 0.022 &
			0.352 & 0.038 & 0.339 & 0.009 &
			0.499 & 0.040 & 0.469 & 0.027 \\ \cline{1-13}
			RegDVAE & 0.469 & 0.014 & 0.430 & 0.020 &
			0.377 & 0.020 & 0.466 & 0.036 &
			0.587 & 0.005 & 0.451 & 0.005 \\
			RegDVAE+D & \textbf{0.499} & 0.022 & \textbf{0.497} & 0.013 &
			\textbf{0.432 }& 0.028 & \textbf{0.589} & 0.071 &
			\textbf{0.665} & 0.022 & \textbf{0.562} & 0.038 \\
            \bottomrule

        \end{tabular}
        }
\vspace{-0.1in}
  \caption{Comparison of prediction results based on encoder using NYT122, NYT71, and NYT27 datasets with different KB regularization strategies.  }
\label{table:comparetable}
\vspace{-0.1in}
\end{table*}

    \begin{figure*}[t]
    \vspace{-0.in}
    \center
    \subfigure[Parameter sensitivity analysis at $\beta$.] {\includegraphics[width=0.32\textwidth]{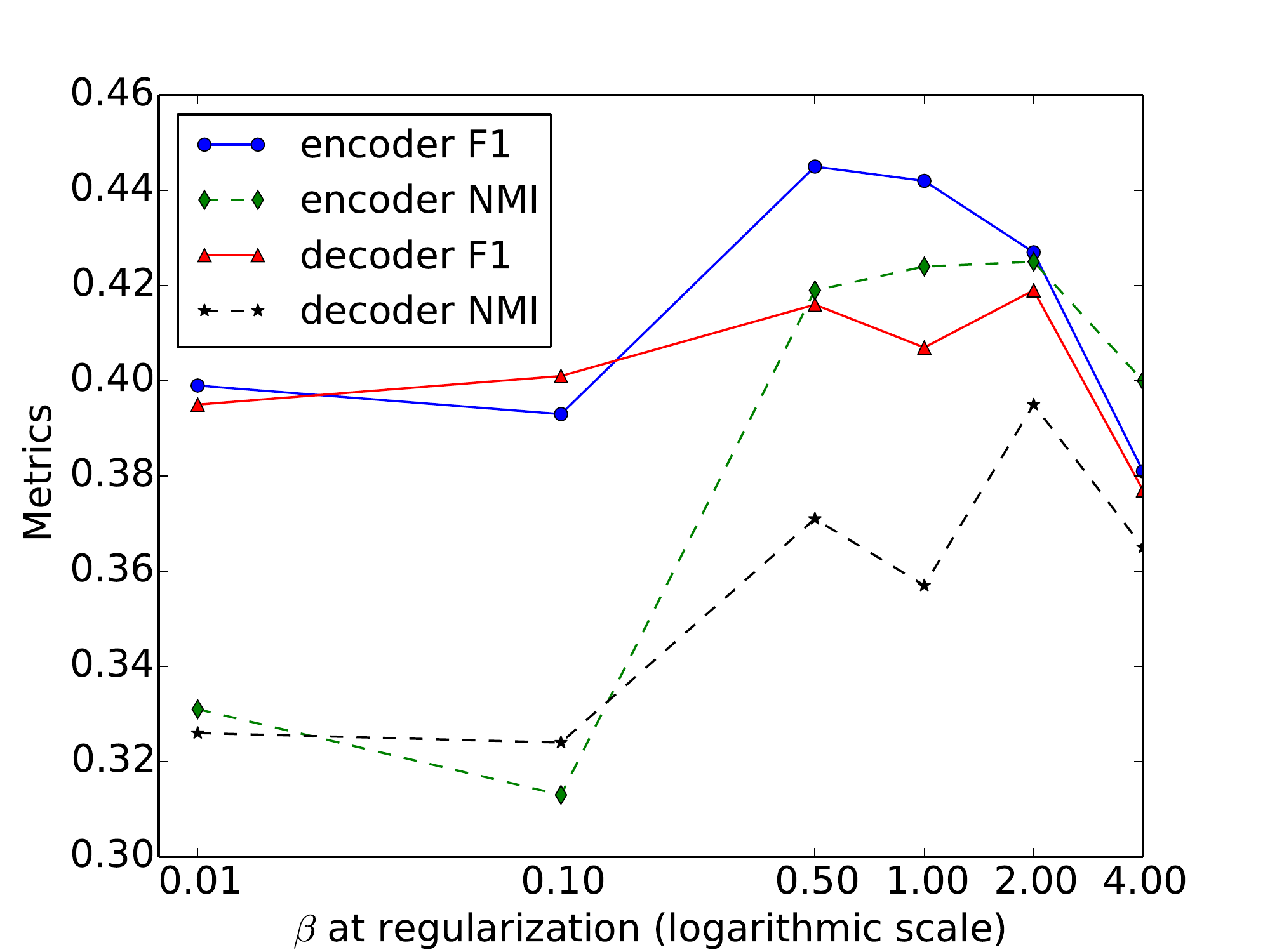}\label{fig:beta}
    }
    \subfigure[Parameter sensitivity analysis at $\gamma$.] {\includegraphics[width=0.32\textwidth]{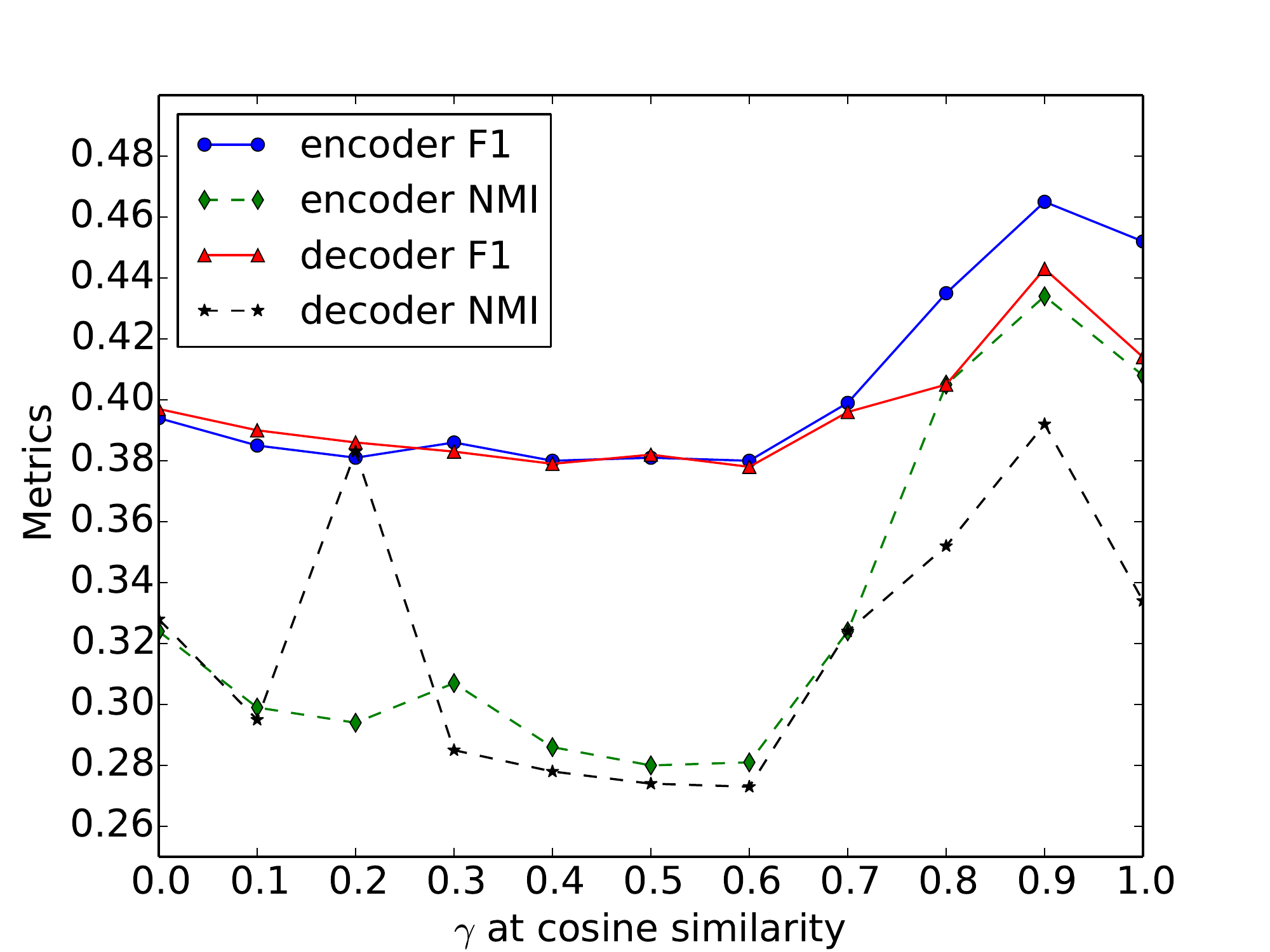}\label{fig:gamma}
    }
    \subfigure[Relation overlap ratios.] {\includegraphics[width=0.32\textwidth]{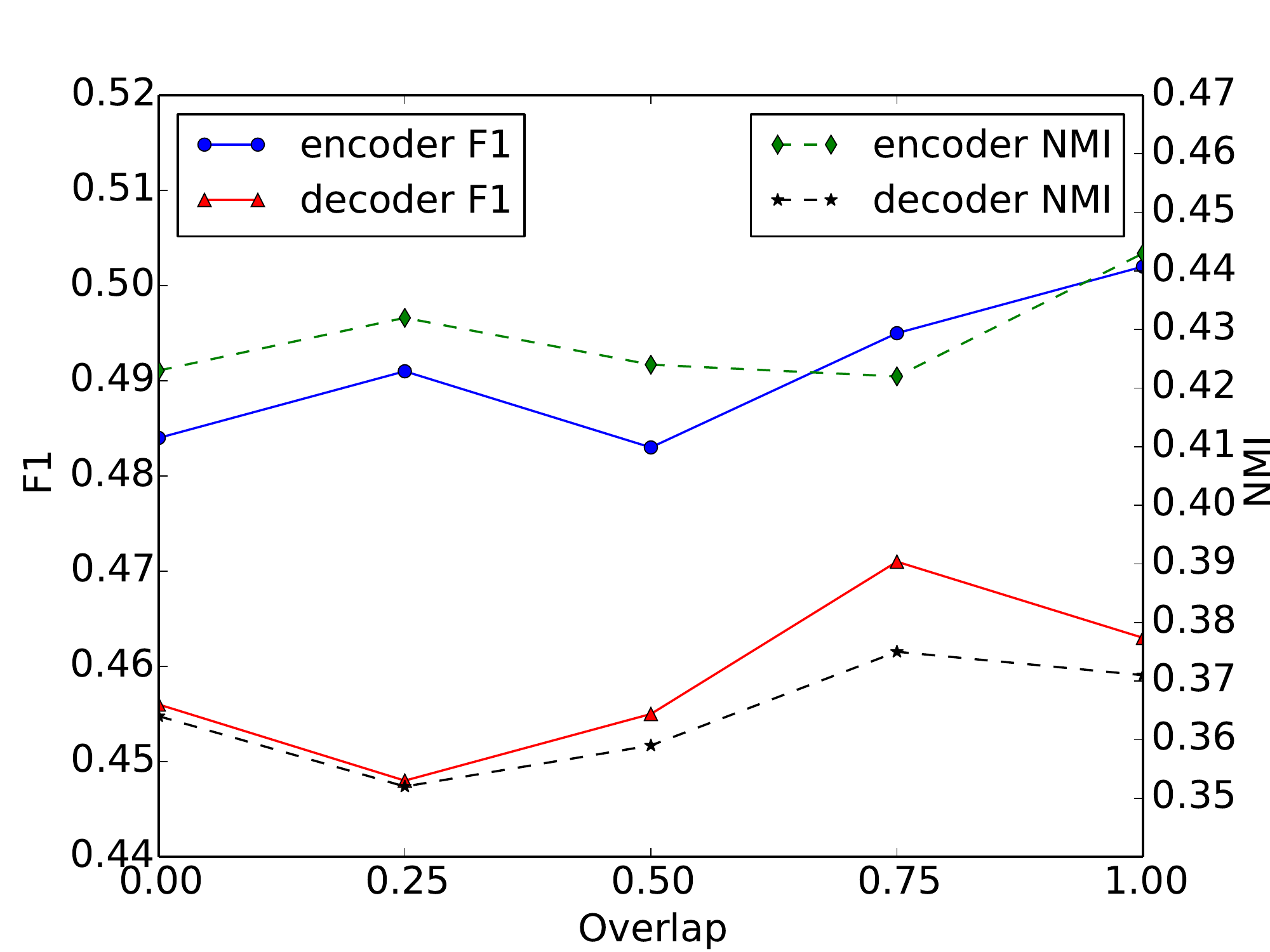}\label{fig:ct}
    }
    \vspace{-0.1in}
        \caption{Comparison results on NYT122 with different parameters and relation overlaps. The predictions are based on either the encoder or the decoder.}
    \vspace{-0.1in}
    \end{figure*}
	\paragraph{Hyper-parameter Sensitivity.}
    We have three hyper-parameters in our algorithm: $\alpha_0$ for the regularization of encoder entropy, $\beta$ for the regularization with our constraints, and $\gamma$ for the threshold of KB based cosine similarities. Here, we test $\beta$ and $\gamma$, since the sensitivity result of $\alpha_0$ is the same as the original DVAE work~\cite{MarcheggianiT16}.
    The sensitivity of $\beta$ is shown in Figure~\ref{fig:beta}. The results are good in a wide range from $\beta=0.5$ to $\beta=2$.
    The sensitivity of $\gamma$ is shown in Figure~\ref{fig:gamma}.
    It reveals some interesting patterns. At the beginning when $\gamma$ is small, it hurts the performance. After $\gamma$ getting greater than 0.7, it improves the performance, which means that only very similar relations indicated by KB embeddings are useful relations as constraints.
	In addition, $\gamma=1$ (meaning only finding identical relations)  is worse than $\gamma=0.9$, which means we indeed find some relations in our KB so that different triplets will be constrained.

	\paragraph{KB Relation Overlap.}
	Although we assume that there is no overlapped relation between the KB and the training text corpus, in practice, we may find a lot of applications that the relations are partially observed in KB.
    Thus, we also test a setting when the KB has different proportions of overlapped relations with training text corpus.
    In this case, we train different KB embeddings for different percentages of overlapped relations, and then apply the embeddings into the constraints.
    The results are shown in Figure~\ref{fig:ct}.
    As we can see, in general, more overlapped relations will result in better performance.
    The best number can be better than the number without overlapped relation by about two points.
    This again verifies that the KB embedding is very robust and represent the semantic meanings of entities even with part of the relations observed~\cite{BordesUGWY13}.

    \begin{table}[t]
      \centering
      {\small
        \begin{tabular}{p{4cm}|c|c}
          \toprule
          Contextual Sentence &  Cluster & Similarity \\
          \midrule
          \ldots \textbf{Spain} will become the third country in \textbf{Europe}\ldots
                              &  12 & \multirow{2}{*}{0.926}\\
          \textbf{Portugal}, with all that talent, goes home to \textbf{Europe}\ldots
                              & 12 & \\
          \midrule
          \textbf{Brazil}, \textbf{Latin America} 's largest economy    \ldots
& 12 & \multirow{2}{*}{0.916}\\
          \ldots \textbf{Argentina} was perhaps the most expensive country in \textbf{Latin America} for tourists.\ldots & 12 & \\
          \bottomrule
        \end{tabular}
      }
     \vspace{-0.1in}
      \caption{Examples for relation: /location/contained\_by.}
      \label{table:examples}
       \vspace{-0.25in}
    \end{table}

\paragraph{Case Study.} We also show some examples of entity pair similarities in Table~\ref{table:examples}. From the Table we can see that our target relation cluster is {\it /location/contained\_by}. In the first example, the similarity between entity pairs (\texttt{Spain, Europe}) and (\texttt{Portugal, Europe}) are high, which indicates the same cluster of pairs of sentences. The same constraint is applied in the second example, although there's no direct connection between (\texttt{Brazil, Latin America}), (\texttt{Argentina, Latin America}).


\section{Related Work} \label{section:previous}
\vspace{-0.1in}
\paragraph{Supervised and Distantly Supervised Relation Extraction.}
Traditional supervised relation extraction focuses on a limited number of relations~\cite{RothY02,Kambhatla04,ChanR10}.
Distant supervision uses KBs to obtain a lot of automatically annotated data~\cite{mintz2009distant,RiedelYM10,HoffmannZLZW11,SurdeanuTNM12,XuHZG13,DBLP:conf/acl/ZhangZZYCS13,ZengLC015,LinSLLS16,ZengLLS17}.
There are two important assumptions behind these models, namely multi-instance learning \cite{RiedelYM10} and multi-instance multi-label learning \cite{HoffmannZLZW11,SurdeanuTNM12}.
Our setting is similar to multi-instance learning but we assume there is no overlapped relation between KB and training text corpus.
Universal schema \cite{RiedelYMM13,VergaBSRM16,ToutanovaCPPCG15, DBLP:conf/eacl/McCallumNV17} can also exploit KB to help extract relations.
It needs a lot of entity pairs in text to co-occur with KB triplets, which is under the same setting with distant supervision. Those surface patterns are pre-extracted and shown in the training phase, which makes it also a weakly supervised learning method.

\paragraph{Unsupervised Relation Extraction.} 
Open Domain Information Extraction (Open-IE) assumes that every relation expression can represent a unique relation~\cite{knowitall,BankoCSBE07,FaderSE11,MausamSSBE12,XuKQGB13,AngeliPM15}.
On the other hand, relation clustering approaches group all the related relation expressions to represent a relation~\cite{LinP01a,MohamedHM11,TakamatsuSN11,YaoHRM11,YaoRM12,NakasholeWS12,NakasholeWS12b,MarcheggianiT16}.
Our setting is based on \cite{MarcheggianiT16} but we also introduce KB as a different kind of weak and indirect supervision.

\paragraph{Knowledge Base Representation.}
Embedding based knowledge base representation learning methods \cite{BordesUGWY13, WangZFC14, LinLLSRL15, TrouillonWRGB16} represent entities and relations as vectors, denoted as $\mathbf{e}$ and $\mathbf{C}_r$ respectively such that for a distances function $f$, the value $f(\mathbf{e}_1, \mathbf{C}_r, \mathbf{e}_2)$ is maximized for all $(e_1, r, e_2)$ facts. Among all these methods, TransE model has a favorable property that the translation operation can be easily recovered by entity vectors $(\mathbf{r}_{1,2} =\mathbf{e}_1-\mathbf{e}_2)$. With its simplicity and high performance, TransE is enough for demonstration. Though our method is not restricted to the representation form of KB, we leave it for future evaluation.

Constraints can be made more explainable by paths finding. For instance, the Path Ranking Algorithm (PRA) \cite{LaoC10,LaoMC11} uses random walk to perform multi-hop reasoning based on logic rules. Later on,
reinforcement Learning \cite{ToutanovaCPPCG15, XiongHW17, Das2017, Chen2018VKGR} is used to search for paths more effectively. Though heuristics are used to further reduce the number of mined relations, it is still very costly to find the paths for KB with hundreds of relations, if not impossible.

\paragraph{Constraint Modeling.}
Originated from semi-supervised learning~\cite{ChaSchZie06}, must-link and cannot-link modeling has been well studied in machine learning community~\cite{WagstaffCRS01,BasuBM04,Basu08}.
Such constraints were usually generated based on the ground truth labels of data.
For document clustering, word constraints constructed based on WordNet similarities have been applied~\cite{DBLP:journals/tkde/SongPLWZQ13} and entity constraints based on entity types in an external KB have been used~\cite{DBLP:conf/kdd/WangSERZH15,DBLP:journals/tkdd/WangSRZH16}, both being considered as a kind of indirect supervision based on side information.
For triplet relation clustering, relation surface similarity and entity type constraints have been explored~\cite{DBLP:conf/ijcai/WangSRWHJZ15}.
However the above constraints are applied to a particular form of models, co-clustering models.
Compared to existing approaches, our constraints are constructed based on more recently developed KB embeddings, which is more flexible and easy to incorporate into different models.

In natural language processing community, constraints based on background knowledge are also well studied.
For example, constrained conditional models (CCM)~\cite{ChangRR12} provides a very flexible framework to decouple learning and inference, where in the inference step, background knowledge can be incorporated as an ILP (integer linear programming) problem. 
Posterior regularization (PR) \cite{ganchev2010posterior} generalizes this idea so that it uses a joint learning and inference framework to incorporate the background knowledge.
Both CCM and PR have many applications including the application to relation extraction~\cite{ChanR10,ChenBNB11}.
Compared to these existing approaches, our constraints are derived from the general-purpose KB, which is quite different from their way of manually crafting some background knowledge as declarative rules.

It is very interesting that we are similar to the PR framework. Since we use a DVAE framework as the base algorithm, there is no traditional E-step and M-step in the variational inference. Instead, only $q$ and $p$ probabilities parameterized by neural networks are updated.
In our framework, we can add constraints to either $q$ or $p$ probabilities (applying to $p$ needs modification of normalization).
It is the same that we draw a biased learning process when estimating the posteriors as PR does.


\section{Conclusion} \label{sec:conclusion}
In this paper, we propose a new relation discovery setting where there is no overlapped relations between the training text corpus and the KB.
	We propose a new learning framework of KB regularization which uses must-link and cannot-link constraints derived based on similarities in the KB embedding space.
    Our method improves the results over all baseline models without harming the scalability. 
    We believe this framework is as flexible as other constraint models to be applied to many applications when we think the semantics of entities and relations provided by the KB is useful. 
    
\section*{Acknowledgments}
This paper was supported by the Early Career Scheme (ECS, No. 26206717) from Research Grants Council in Hong Kong. We thank Intel Corporation for supporting our deep learning related research. We also thank the anonymous reviewers for their valuable comments and suggestions that help improve the quality of this manuscript.

\bibliography{yanbib}
\bibliographystyle{styles/acl_natbib}

\end{document}